\documentclass[letterpaper, 10 pt, conference]{ieeeconf}  % Comment this line out if you need a4paper

\IEEEoverridecommandlockouts       % This command is only needed if  you want to use the \thanks command
\usepackage{style}

\title{\LARGE \bf
Dynamic Tube MPC: Learning Tube Dynamics with Massively Parallel Simulation for Robust Safety in Practice
}

\author{William D. Compton$^1$, Noel Csomay-Shanklin$^1$, Cole Johnson$^2$, Aaron D. Ames$^1$ % <-this % stops a space
\thanks{$^1$The authors are with the Department of Computing and Mathematical Sciences, California Institute of Technology, Pasadena, CA.}
\thanks{$^2$ The author is with the College of Computing, Georgia Institute of Technology, Atlanta, GA.}
\thanks{This research is supported by Technology Innovation Institute (TII).}
}

\begin{document}

\maketitle
\thispagestyle{empty}
\pagestyle{empty}

%%%%%%%%%%%%%%%%%%%%%%%%%%%%%%%%%%%%%%%%%%%%%%%%%%%%%%%%%%%%%%%%%%%%%%%%%%%%%%%%
\begin{abstract}

Safe navigation of cluttered environments is a critical challenge in robotics. It is typically approached by separating the planning and tracking problems, with planning executed on a reduced order model to generate reference trajectories, and control techniques used to track these trajectories on the full order dynamics. Inevitable tracking error necessitates robustification of the nominal plan to ensure safety; in many cases, this is accomplished via worst-case bounding, which ignores the fact that some trajectories of the planning model may be easier to track than others. In this work, we present a novel method leveraging massively parallel simulation to learn a dynamic tube representation, which characterizes tracking performance as a function of actions taken by the planning model. Planning model trajectories are then optimized such that the dynamic tube lies in the free space, allowing a balance between performance and safety to be traded off in real time. The resulting \emph{Dynamic Tube MPC} is applied to the 3D hopping robot ARCHER, enabling agile and performant navigation of cluttered environments, and safe collision-free traversal of narrow corridors. 

\end{abstract}

%%%%%%%%%%%%%%%%%%%%%%%%%%%%%%%%%%%%%%%%%%%%%%%%%%%%%%%%%%%%%%%%%%%%%%%%%%%%%%%%
\section{Introduction}

Safe autonomous navigation of cluttered environments is a challenging problem for robotic systems. 
Issues including non-convexity of the traversible environment, complexities of a robot's full-order dynamics, and the desire to simultaneously achieve dynamic and safe behaviors during execution motivate hierarchical solutions to this problem. 
Typical approaches decompose the problem into a \emph{planning layer}, operating on a simplified model to find a collision-free (safe) path through the environment, and a \emph{tracking layer}, which takes in the plan and produces control inputs to track it as accurately as possible on the full order system. 

When decomposed in this way, the primary issue of interest becomes certifying how well the tracking controller can track plans, as error between the two systems can cause unsafe behaviors, i.e., collisions, if not accounted for properly. 
Most commonly, robotic practitioners will simply add a conservative, ad-hoc safety margin to the planning system; however significant work has been performed to develop theoretically grounded, less conservative methods. 
By setting up a pursuit-evasion game between the tracking and planning models, FasTrack leverages numerical solutions to the HJB PDE to identify a tracking error bound, which upper bounds the deviation between the two models \cite{Chen2021}. 
For systems of specific structure (for instance, feedback linearizability), ISS Lyapunov functions can be identified, whose level-sets are forward invariant and thus give a tracking error bound \cite{Csomay-S2022}, and input-to-state safe (ISSf) control barrier functions (CBFs) can be utilized with the reduced order dynamics to ensure safety \cite{Molnar2022} (see \cite{Cohen_2024,wabersich2023data} for recent surveys). In the case where a Lyapunov function cannot be identified analytically, a space of functions can be parameterized and searched, for instance via sum-of-squares programming \cite{Tobenkin2010} or learning \cite{dawson2022safe, Bansal2020}. This problem has also been treated in the language of simulation functions \cite{Girard2009, Kurtz2020}, and contraction theory \cite{Zhao2021}. 
\begin{figure}[t!]
    \centering
    \includegraphics[width=\columnwidth]{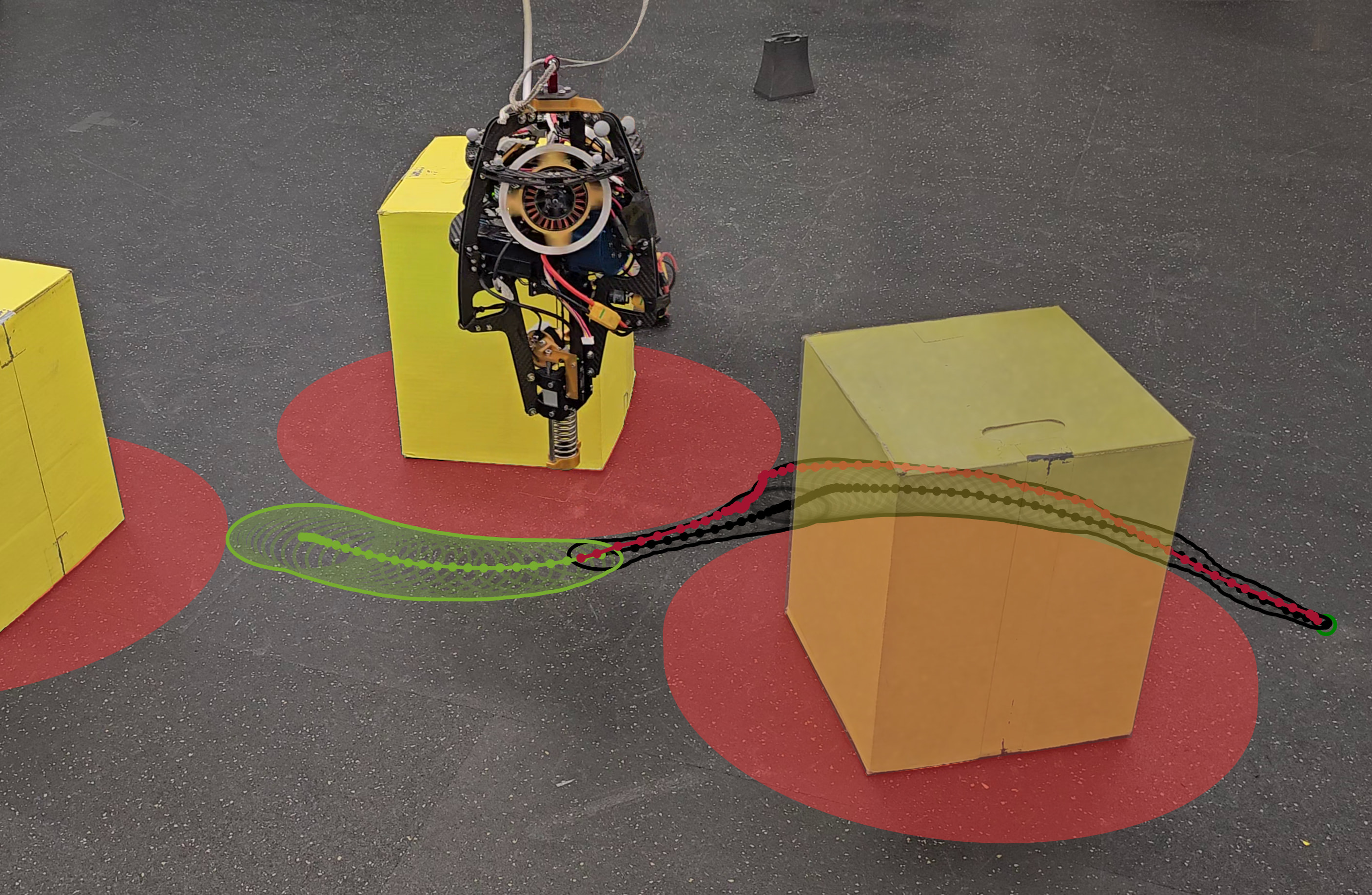}
    \caption{ARCHER plans a collision-free path through a cluttered environment by jointly optimizing a path on a reduced order model and a dynamic tube, whose dynamics are learned from simulation data. The obstacles are tightly approximated by circles and buffered by the radius of the robot.}
    \label{fig:hero}
    \vspace{-6mm}
\end{figure}

Once a tracking invariant has been established, the navigation problem is simplified, and now requires finding a path for the planning model which, when buffered by the tracking invariant, is collision free. 
Tube Model Predictive Control (Tube MPC) solves this problem by minimizing a cost function while ensuring the tube-buffered trajectory lies in the free space \cite{Langson2004, Sieber2021}.
While computationally intensive, Tube MPC has been implemented in real-time to control several hardware systems, including \cite{do:hal-04620816, MICHEL2019112}.
A notable drawback of Tube MPC formulations is that the fixed, worst case tubes can be quite conservative; to mitigate this, Planning Fast and Slow introduces a meta-planner to select what planning model (and thus tube size) to use depending on the environment \cite{Frido2017}. 
Dynamic Tube MPC offers further flexibility by allowing tube sizes to vary with control decisions, as demonstrated through implementations with sliding mode control \cite{Lopez2019} and learned tube dynamics \cite{Fan2020}.  Alternatively, stochastic methods can be used to characterize dynamics uncertainty. 
Methods either rely on the mean and variance \cite{hakobyan2019, cosner2024generative}, or the tail probabilities \cite{koenker1978regression, Bradford2019, Bujarbaruah2019} to plan paths which are highly likely to be collision free. 
Building on \cite{Fan2020}'s use of quantile regression for tube learning, our approach significantly increases the available training data, incorporates error history information for more accurate tube prediction, and achieves real-time performance on a complex hardware platform. 

A recent trend in robotics has been to leverage massively parallel simulation to learn desired behaviors through significant environmental interaction. Reinforcement learning, at the center of this approach, has seen significant success, including work in drones \cite{Kaufmann2023}, legged robotics \cite{Rudin2021}, and bipeds \cite{Liao2024}.
Recent work has even used parallel simulation for online planning \cite{pezzato2023samplingbased}.
While learning has been used in the past to generate certificates for dynamical systems, including Lyapunov/barrier functions, \cite{dawson2022safe, Chang2020, Mathiesen2023}, reachable sets \cite{Bansal2020}, and stabilizing policies, \cite{Compton2024}, there is an opportunity to extend these learning based certification schemes to significantly more complex and high dimensional environments by leveraging massively parallel simulation. 

In this paper, we focus on the critical insight that for many systems, tracking performance varies with the signal being tracked. Very aggressive trajectories are difficult to track, resulting in large error bounds, while conservative trajectories can be tracked more closely. Informed by this concept, we learn a dynamic error tube that associates a planning model trajectory with a corresponding tube, within which the tracking model is expected to remain while following the trajectory. We condition the tube on a history of tracking errors, and demonstrate significant tube accuracy gains with increasing history length. Once the tube dynamics have been learned, we formulate a \emph{Dynamic Tube MPC} problem, which optimizes trajectories of the planning model such that the dynamic tube remains in the free space.

Dynamic Tube MPC is deployed to control the 3D hopping robot ARCHER to safely navigate cluttered environments. We demonstrate problems in which fixed-size tube MPC leads to either infeasibility or highly conservative behavior, and demonstrate that our Dynamic Tube MPC achieves both performance and safety. \cref{fig:hero} shows ARCHER running Dynamic Tube MPC in real time to dynamically and safely navigate a cluttered environment.

\section{Preliminaries}
In this section, we introduce the mathematical concepts and nomenclature to be used throughout the paper. Discrete time variables will be indexed via subscript (i.e. $\b x_k$), and sequences will be indicated via colon notation, $\b x_{i:j} = \{\b x_i, \b x_{i+1}, \hdots, \b x_{j-1}\}$. We begin by defining the tracking and planning models, which form the foundation of our approach.

\subsection{Tracking and Planning Models}
Considering a system with $n_x$ states and $m_x$ inputs, with discrete dynamics governed by:
\begin{equation}
    \b x_{k+1} = \b f_{\b x}(\b x_k ,\b u_k)
\end{equation}
where $\b x_k \in \cal{X} \triangleq \R^{n_x}$ is the system state, $\b u_k \in \mathcal{U} \triangleq \R^{m_x}$ the system input, $\b f_{\b x}: \cal{X} \times \cal{U} \to \cal{X}$ the tracking dynamics, and $k \in \Z$ indexes time step.
Note that while many robotic systems are naturally described by differential equations, we consider discrete time systems as they will be amenable to both the learning problem and the planning problem.
For our purposes, these can be the discretized version of continuous dynamics (for instance, as discretized by a simulator). 

To facilitate planning, we also define a planning model, with $n_z < n_x$ states and $m_z < m_x$ inputs,
\begin{equation}
    \b z_{k+1} = \b f_{\b z}(\b z_k ,\b v_k)
\end{equation}
where $\b z_k \in \cal{Z} \triangleq \R^{n_z}$ is the planning state, $\b v_k \in \mathcal{V} \subset \R^{m_z}$ the planning input, and $\b f_{\b z}: \cal{Z} \times \cal{V} \to \cal{Z}$ the planning dynamics. 
Additionally, we consider a tracking controller $\b k : \cal{X} \times \cal{Z} \times \cal{V} \to \cal{U}$, which selects control inputs for the full order system to track the planning system. 

The planning dynamics should include all states relevant to the planning problem; for instance, in a collision avoidance problem, the planning model should contain at least the position states, as these are relevant to the constraints. To this end, we define a projection operation $\b \Pi_{\b z}: \cal{X} \to \cal{Z}$, which takes the tracking system state and projects it onto the planning model. From this, we can define the planning model error $e : \cal{X} \times \cal{Z} \to \R_{\geq 0}$:
\begin{equation} \label{eqn:err}
    e_k(\b x_k, \b z_k) = \|\b z_k - \b \Pi_{\b z}(\b x_k) \|
\end{equation}
More general definitions of the error can be taken; ours parameterizes the error as a scalar, and leads to spherical tubes for ease of planning (see \cref{sub:methods_dtmpc}). Other parameterizations, for instance hypercubes or ellipsoids, lead to different structures of the planning problem.

Finally, we introduce the tube size, $w_k \in \R_{\geq 0}$. 
A dynamic tube is a sequence of tube sizes $\{w_j\}$; we define the tube to be \emph{correct} when the closed loop execution of the full order model tracking the planning model yields $w_k \geq e_k$, i.e. the tracking system remains with the tube centered about the planning model. 
Ultimately, we will be concerned with learning tube dynamics, which will predict future tube sizes from a history of relevant data. 

\subsection{Nominal MPC for Planning}
In the absence of tracking error, we can consider a nominal planning problem. Consider a environment with free space (collision-free subset of the planning model state space) denoted $\cal{C} \subset \cal{Z}$. For a given initial condition $\b z_i$, goal state $\b z_g$, and cost functional $J:\cal{Z}^{N+1} \times \cal{V}^N \to \R_{\geq 0}$, the nominal MPC problem is then given:
\begin{mini!}[1]
{\b v_{(\cdot)}, \b z_{(\cdot)}}{J(\b z_{(\cdot)}, \b v_{(\cdot)}) \label{eq:nom_mpc_obj}}
{\label{eqn:nom_mpc}}{}
\addConstraint{\b z_{j+1}}{=\b f_{\b z}(\b z_j, \b v_j) \quad \label{eqn:nom_dyn}}{j \in [0, N-1]}
\addConstraint{\b z_0}{= \b z_{i} \label{eqn:nom_ic}}
\addConstraint{\b v_j}{\in \cal{V} \quad \label{eqn:nom_input}}{j \in [0, N-1]}
\addConstraint{\b z_j}{\in \cal{C} \quad \label{eqn:nom_safe}}{j \in [0, N]}
\end{mini!}
where $N \in \Z_{>0}$ is the length of the planning horizon. Under appropriate selection of the cost function, solving this problem in closed loop by applying the first input in the solution trajectory and then resolving the MPC problem will result in stability of the closed loop system to the goal \cite{Williams2001}. 

\subsection{Application to ARCHER}
The ARCHER 3D hopping robot, shown in \cref{fig:hero}, is the primary hardware focus of this paper. Stabilizing controllers for ARCHER have been previously studied \cite{csomay2024robust, Csomay2023}.
The system contains $n_x=16$ states $\b x = [\b p^\top \; \b q^\top \; \b v^\top \; \b \omega^\top \; \dot{\b \theta}{}^\top ]^\top$, where $\b p \in \R^3$ is the global position, $\b q \in \mathbb{S}^3$ a unit quaternion defining orientation, $\b v \in \R^3$ the linear velocity, $\b \omega \in \mathfrak{s}^3$ the angular velocity, and $\dot{\b \theta} \in \R^3$ the velocities of the flywheels. 
The system takes $m_x=3$ inputs, the torques applied to each of the flywheels. The foot subsystem is autonomous, and not considered in the design of our controllers; the foot spring is compressed to a preset distance during the flight phase, and released during the ground phase to maintain approximately constant hop height.
More details regarding the hardware can be found here \cite{Csomay2023}. The system dynamics $\b f_{\b x}$ will simply be steps of the robot in a given simulator. 

In this work we will plan trajectories for the hopper using a two dimensional single integrator with $\b z_k \in \R^2$ the positions and $\b v_k \in \mathcal{V} \subset \R^2$ the velocities, which will be bounded to lie in a symmetric range, $-\bar{v} \leq \b v_k \leq \bar{v}$, where $\bar{v} \in \R_{\geq 0}$ is the velocity bound. 
The projection operator will simply return the $x$ and $y$ positions of the hopper, $\b \Pi_{\b z}(\b x_k) = [p_{x, k} \; p_{y,k}]^\top$. 

The tracking controller used for ARCHER in this work will be a Raibert Heuristic \cite{raibert1984experiments}. Given a set of gains $k_p, k_d, k_f \geq 0$, and clipping parameters, $c_p, c_v, c_f, c_a \geq 0$ we compute the desired roll, $\phi_d$, and pitch, $\gamma_d$, as:
\begin{subequations}
\begin{align}
    \b e_p &= \text{clip}\left(\b z - \b \Pi_{\b z}(\b x)  ,c_p\right) \\
    \b e_v &= \text{clip}\left(-\b \Pi_{\b v}(\b x) ,c_v\right) \\
    \b e_f &= \text{clip}\left(\b v ,c_f\right)  \\
    [\phi_d, \; -\gamma_d]^\top &= \text{clip}\left(-k_p \b e_p -k_d \b e_v + k_f \b e_f, c_a\right)
\end{align}
\end{subequations}
where $\b \Pi_{\b v}: \cal{X} \to \R^2$ projects the hopper state to its $x$ and $y$ velocity coordinates.
The clipping operations prevent large errors from requesting too large of orientations, which lead to falling.
Finally, geometrically consistent PD control is used to track the desired orientation as outlined in \cite{Csomay2023}. 

\begin{figure*}  % TODO: placement of this figure
    \centering
    \includegraphics[width=\textwidth]{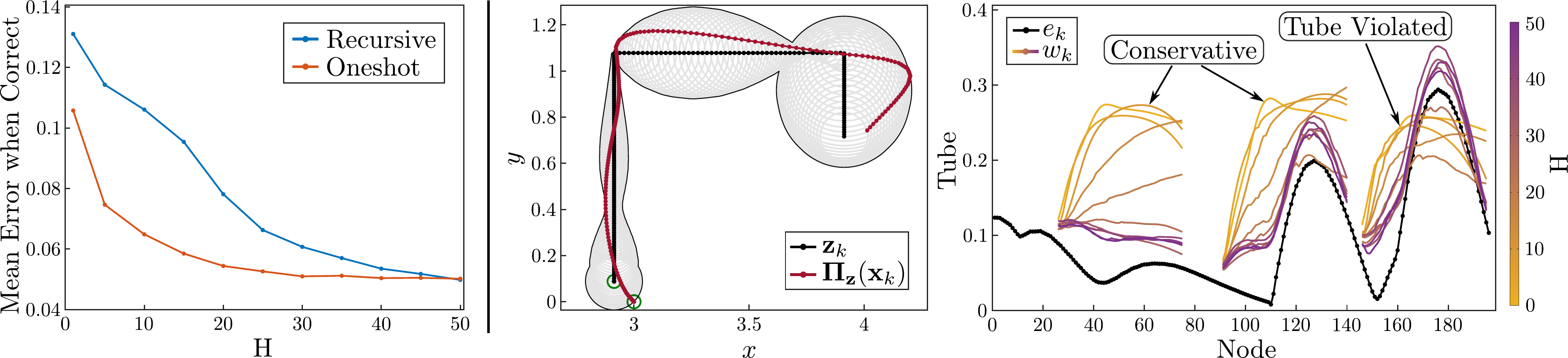}
    \caption{Analysis of the impact of history, $H$, on tube prediction accuracy; tube dynamics are trained on the hopper dynamics for several values of $H$. (Left) Evaluation on a 20\% holdout from training data, Mean Error when Correct is plotted against history length. (Middle) a portion of a square trajectory is executed in IsaacGym on ARCHER. (Right) The error signal, $e_k$, is plotted in black, with the tube predictions, $w_k$, plotted at three points in time, $k=25,75, 145$ over a horizon of $N=50$ nodes (right pane). We see short horizon models being both overly conservative and violating the error tube.}
    \label{fig:hist_comp}
    \vspace{-6mm}
\end{figure*}

\section{Methods}

Our approach consists of two primary components: learning a dynamic representation of an error tube that characterizes the how tracking error bounds depend upon trajectories of the planning model, and employing Dynamic Tube MPC to plan collision-free paths within these bounds. This section details each of these components. 

\subsection{Learning Tube Dynamics} \label{sub:tube_learning}
The primary goal of this work is to characterize how the actions taken by the planning model effect the size of the tracking invariant established by the tracking controller. To this end, consider the following problem.
Take any feasible trajectory of the planning model, covering a length of at least $H + N$ time steps, where $H \in \Z_{\geq 0}$ is the history considered by the model, and $N \in \Z_{> 0}$ the horizon for the planning problem). 
Assume the full order system has been tracking the planning model for at least $H$ time steps. Then at the current time $k$, we can compute the a history of tracking error as $e_{k-H:k}$ by applying \eqref{eqn:err} at each of the time steps.
Now, given the state and input trajectory of the planning model, $\{\b z_j\}_{j=k}^{k+N}, \{\b v_j\}_{j=k}^{k+N}$, we would like to predict the tube, $\{w_j\}_{j=k}^{k+N}$, such that $w_j \geq e_j$ at each point along the trajectory when executing the tracking controller in closed loop. We consider two methods of prediction:
\begin{itemize}
    \item One-Shot Tube Dynamics:
    \begin{equation}
        w_{k:k+N} = f_{w,os}(e_{k-H:k}, \b z_{k-H:k+N}, \b v_{k-H:k+N})
    \end{equation}
    This predicts the tube size over the horizon as a single function of the error history and planning trajectory.
    \item Recursive Tube Dynamics: 
    \begin{align}
        w_{j+1} &= f_{w,rec}(\Tilde{e}_{j-H:j}, \b z_{j-H:j}, \b v_{j-H, j}, j) \label{eqn:rec_tube} \\
        \Tilde{e}_{j-H:j} &= \begin{cases}
            [e_{j-H:k} \; w_{k:k+H-j}] & j < k+H \\
            w_{j-H:j} & j \geq k+H
        \end{cases} \nonumber
    \end{align}
    This is applied recursively from $j=k$ to $j=k+N$, predicting the next tube size as a function of the history of tube sizes and planning trajectory. Critically, for indices of $\Tilde{e}_{j-H:j}$ up to $k$, we will use the error history, while after $k$ we use the tube prediction. For this reason, $j$ is also passed to the function, as this will help distinguish how much of $\Tilde{e}$ is error history as opposed to tube predictions. 
\end{itemize}

% To learn the tube dynamics, we utilize massively parallel simulation in IsaacGym \cite{isaacgym} to collect a large dataset of the form $\cal{D} = \{\b z_{0:\Bar{N}+1}, \b v_{0:\Bar{N}},  \b \Pi_{\b z}(\b x_{0:\Bar{N}+1})\}$, with $\Bar{N}$ the length of each trajectory in the dataset.
To learn the tube dynamics, we utilize massively parallel simulation to collect a large dataset of the form $\cal{D} = \{\b z_{0:\Bar{N}+1}, \b v_{0:\Bar{N}},  \b \Pi_{\b z}(\b x_{0:\Bar{N}+1})\}$, with $\Bar{N}$ the length of each trajectory in the dataset.
The inputs to the planning model are randomly generated\footnote{The distribution of trajectories matters when deploying the tube model in an MPC problem, as discussed in \cref{sec:application_tube_learning}. See code \cite{code} for implementation details.} and the planning model and tracking model initial conditions are sampled from an initial condition sets, $\b z_0 \sim \cal{Z}_0 \subset \cal Z$, and $ \b x_0 \sim \cal{X}_0(\b z_0) \subset \cal{X}$. 

The tube dynamics are represented by a neural network, $f_w^{\b \theta}$, with parameters $\b \theta$. To train the model, as in \cite{Fan2020}, inspired by quantile regression \cite{koenker1978regression}, we use a check loss function:
\begin{subequations}
\begin{align}
    \b w(\b \theta) &= f_w^{\b \theta}(\b e_{k-H:k}, \b z_{k-H:k+N}, \b v_{k-H:k+N}) \label{eqn:fw_eval}\\
    \b e &= \|\b z_{k:k+N} - \b \Pi_{\b z}(\b x_{k:k+N}) \|  \label{eqn:gt_err}\\
    \b r(\b \theta) &= \begin{cases}
        \alpha( \b w(\b \theta) - \b e) & \b w(\b \theta) \geq \b e \\ (1 - \alpha) (\b e - \b w(\b \theta)) & \b w(\b \theta) < \b e
    \end{cases}  \label{eqn:check_loss}\\
    \cal{L}(\b \theta) &= \text{Huber}(\|\b r(\b \theta)\|_1) \label{eqn:huber}
\end{align}    
\end{subequations}
Given an index $k$, we evaluate the loss over the horizon length $N$ on which the model predicts. \eqref{eqn:fw_eval} evaluates the tube model (either recursive or one-shot) over the entire horizon. \eqref{eqn:gt_err} then evaluates the tracking error over the horizon. \eqref{eqn:check_loss} implements the check loss, where the $\b w \geq \b e$ condition holds when the tube prediction is correct, and incentives the tube to be tight around the error (by decreasing tube size). The $\b w < \b e$ conditional holds when the tube is incorrect, and incentivizes increasing the tube size. Since the output of \eqref{eqn:check_loss} is a vector (of length $N$), we apply a Huber loss \cite{huber1992robust} to the L1 penalty for a smooth penalty landscape. 

The choice of the parameter $\alpha$ is directly related to the level of robustness achieved by the tube. Previous work \cite{Fan2020} has demonstrated that minimizing the check loss \eqref{eqn:check_loss} corresponds to finding at tube which is an $\alpha$-quantile representation of the error dynamics; the tube is expected to be correct an $\alpha$ proportion of the time.

% Training is conducted in PyTorch \cite{paszke2017automatic}, using neural networks with two layers of 128 nodes, and Softplus ($\beta = 5$) activations, as these result in smooth networks which are easier than ReLU to optimize through in MPC. When training the recursive models, we stop gradients on the $w$ components of $\Tilde{e}$ from \eqref{eqn:rec_tube}, as these terms depend on $\b \theta$; this stabilizes training by localizing gradient information.

\subsection{Dynamic Tube MPC}\label{sub:methods_dtmpc}

Given a dynamic tube model $f_w^{\b \theta}$, which maps an error history and planning model trajectory to a tube prediction (either via one-shot or recursive application), we can formulate the Dynamic Tube MPC problem as a variant of \eqref{eqn:nom_mpc}:

\vspace{-5mm}
\begin{mini!}[1]
{\b v_{(\cdot)}, \b z_{(\cdot)}}{J(\b z_{(\cdot)}, \b v_{(\cdot)}) \label{eq:tube_mpc_obj}}
{\label{eqn:tube_mpc}}{}
\addConstraint{\b z_{j+1}}{=\b f_{\b z}(\b z_j, \b v_j) \quad \label{eqn:tube_dyn}}{{j \in [0, N-1]}}
\addConstraint{w_{0:N}}{{=f_{w}^{\b \theta}(e_{\,\text{-}H:0}, \b z_{\,\text{-}H:N},  \b v_{\,\text{-}H:N})} \label{eqn:tube_tube}}
\addConstraint{\b z_0}{= \b z_{i} \label{eqn:tube_ic}}
\addConstraint{\b v_j}{\in \cal{V} \quad \label{eqn:tube_input}}{{j \in [0, N-1]}}
\addConstraint{B_{w_j}(\b z_j)}{\in \cal{C} \quad \label{eqn:tube_safe}}{{j \in [0, N]}}
\end{mini!}
where $B_r(\b z)$ is a ball of radius $r$ centered at $\b z$. The tube dynamics are enforced via \eqref{eqn:tube_tube}, and safety is enforced by \eqref{eqn:tube_safe}. Dynamic Tube MPC aims to minimize the given cost function while ensuring that the planned trajectory, when buffered by the error tube, remains within the free space. 

In this work, we will consider circular obstacles, which can be encoded as a single constraint, such that the Dynamic Tube MPC problem can be formulated as a nonlinear program (NLP). Given the $i$'th obstacle, centered at $\b z_{c,i}$ with radius $r_i$, the safety constraint \eqref{eqn:tube_safe} can be expressed:
\begin{equation}
    (\b z_j - \b z_{c,i})^\top (\b z_j - \b z_{c,i}) - (w_j + r_i)^2 \geq 0
\end{equation}
% The resulting NLP will be solved via CasADi \cite{Andersson2019} with the SNOPT plugin \cite{gill2005snopt}. To efficiently encode the neural network representation of the tube dynamics, $f_w^{\b \theta}$, we use the L4CasADi toolbox \cite{salzmann2023neural, salzmann2024l4casadi}. 
For closed loop control, we apply the first input in the Dynamic Tube MPC solution, then recompute the solution, with the next initial condition as the current state of the planning model. Note that this differs from some implementations of Tube MPC, where the projection of the tracking system must be within the tube around the initial condition; we avoid this constraint, as it significantly complicates the interpretation and implementation of the error history, $e_{H:0}$.

\section{Analysis}
A critical differentiating factor between our method and previous ones, i.e. \cite{Fan2020}, is the incorporation of an error history, $\Tilde{e}$ in \eqref{eqn:rec_tube}, in the tube prediction. We analyse the impacts of the this error history on the size of the learned tube. Additionally, we highlight the improvements of Dynamic Tube MPC over classical tube MPC approaches.

\subsection{Effect of Error History on Tube Size}
Including a longer error history leads to reductions in the size of the learned tube. To demonstrate this, we train tube dynamics models with histories ranging from $H=1$ to $H=50$, with $\alpha=0.9$, and predict on a horizon of length $N=50$. On a 20\% held out validation set, all models predict tubes which are correct at least 90\% of the time, matching expectations from the check loss \eqref{eqn:check_loss}. The primary metric used to quantify the size of a tube is Mean Error When Correct (MEC); given the errors, $e_{0:N}$, and tube predictions, $w_{0:N}$, the tube is correct on an index set $I_{c} = \{i \in 0, \hdots, N ~\vert~ w_i \geq e_i \}$. The MEC is then
\begin{equation}
    \text{MEC} = \frac{1}{|I_c|} \sum_{i \in I_c} w_i - e_i
\end{equation}
Note each element of this sum is non-negative, as we sum only over indices where the tube is correct. When the MEC is small, the tube tightly approximates the actual error, while when it is large, the tube over-approximates the error significantly. 
In \cref{fig:hist_comp} (left) we see that increasing $H$ leads to reduction in the MEC. For large $H$, these returns diminish. Additionally, we note that the one-shot models tend to out-preform the recursive models at each history length, with this gap closing as $H$ increases. 

To further demonstrate this idea, we consider a case where the planning model executes part of a square, as shown in \cref{fig:hist_comp} (middle) by the black line. The hopper, when tracking this trajectory in simulation, incurs error, as shown by the magenta line.  In \cref{fig:hist_comp} (right), we see three time steps, where the recursive tube predictions are evaluated, for various history lengths of the model. As an overall trend, we see that the short horizon models tend to be significantly more conservative than the long history models; additionally, the long history models seem to capture the shape of the error trajectory much more precisely. In the third set of tube predictions, we see a case where the short-horizon models under-approximate the resulting error, while the long horizon models retain solid performance.  

The trends in \cref{fig:hist_comp} show a strong incentive to use longer error histories when predicting tube size, ideally $H\geq25$ for this setup on ARCHER; however, performance levels off beyond a certain point. The incorporation of the error history is critical, as it allows the model to discriminate between states which have the same instantaneous error (for instance where the error may be increasing or decreasing, leading to very different tubes). We can draw parallels to ideas from filtering; the history of errors and planning trajectories gives the neural network the necessary information to learn a filter of the salient full order model states, and use these to more accurately predict tube sizes over the horizon. Beyond a certain point, including a longer history does not improve the quality of this implicit filter. 
\begin{figure}[t!]
    \centering
    \includegraphics[width=\columnwidth]{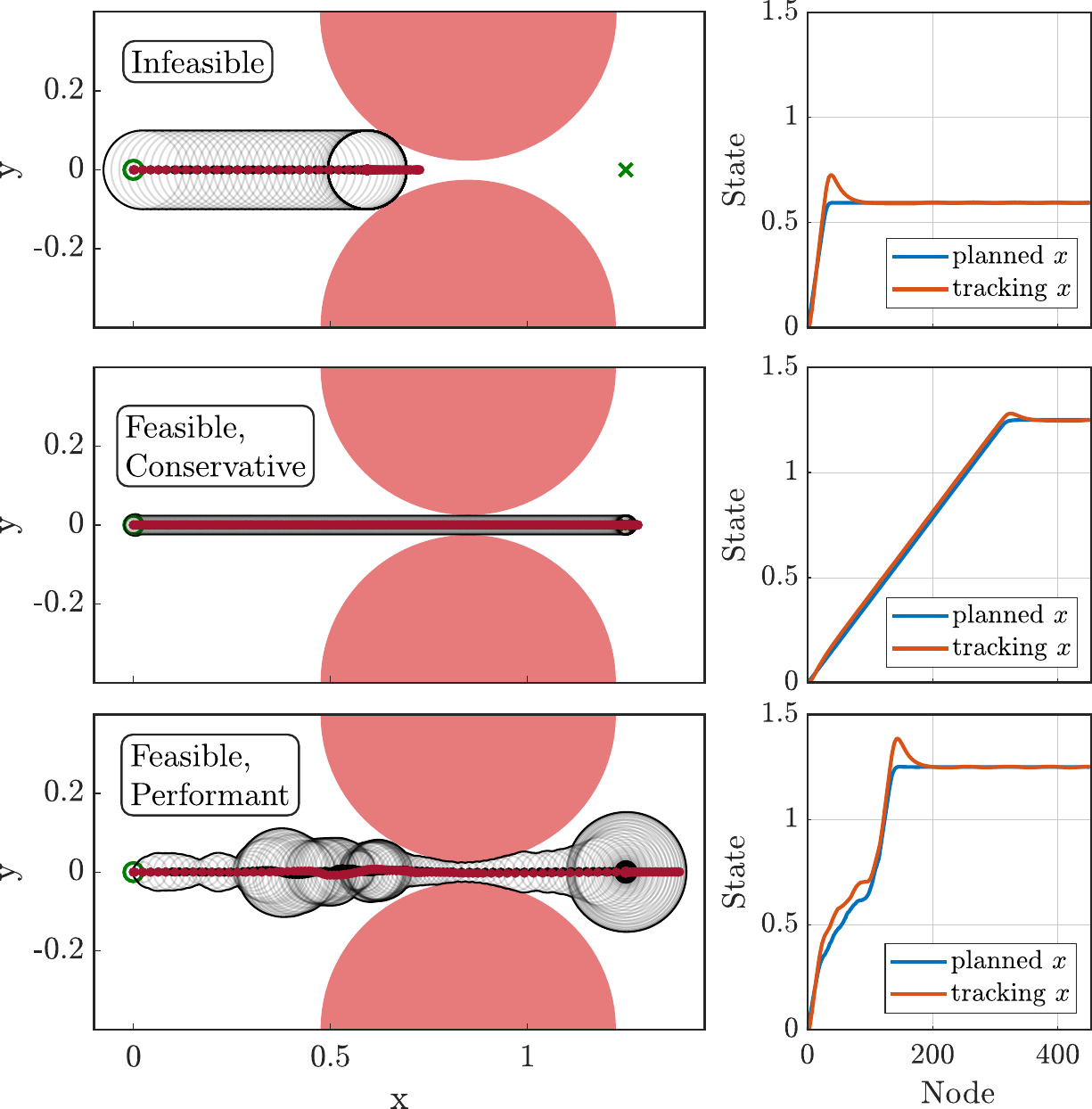}
    \caption{Three Tube MPC variants, on a problem where the hopper must traverse a narrow gap. (Top) Tube MPC, where the tube size is fixed to be the 90\% quantile in a dataset collected with $\bar{v}=0.2$m/s. (Middle) Tube MPC, where the tube size is fixed to the 90\% quantile, in a dataset collected with $\bar{v}=0.04$m/s. (Bottom) Dynamic Tube MPC using recursive tube dynamics, which simultaneously achieves performance and safety.}
    \label{fig:feas}
    \vspace{-6mm}
\end{figure}

\subsection{Dynamic Tube MPC}
When employed in the Dynamic Tube MPC, the learned tube representation allows the optimizer to trade off aggressive behaviors and conservative behaviors when necessary to ensure safety. \cref{fig:feas} demonstrates a critical benefit of Dynamic Tube MPC over a fixed size Tube MPC. A problem is set up for the hopper to traverse a narrow gap; the top and middle panes demonstrate solution of this problem with a fixed tube size, while the bottom pane demonstrates the effectiveness of a recursive dynamic tube, trained on a dataset with $\bar{v}=0.2$m/s. For the top plot, the tube size is the 90\% quantile of the errors in the dataset used to train the tube dynamics (we do this as computation of a true robust invariant is quite difficult, and the quantile allows better comparison to the learned tube, which also uses quantile bounding). For this tube size, the planning model cannot reach the goal. In order to reach the goal with a fixed tube size, we are forced to limit the planning model to $\bar{v}=0.04$m/s; collecting a dataset with these tightened input bounds, the 90\% quantile bound fixed tube allows a very conservative solution to the problem, solving in around 400 nodes (40 seconds). The dynamic tube MPC travels quite quickly when away from the obstacles (note the slope of the state plot), while slowing to a speed similar to the conservative solution while between the obstacles. This allows a much more aggressive trajectory, solving the problem in only 200 Nodes (20 seconds). The ability to dynamically vary the aggressiveness of the planning model allows the system to balance performance with safety, and generates robust and performant behaviors difficult to achieve via other methods. 

\begin{figure*}  % TODO: placement of this figure
    \centering
    \includegraphics[width=\textwidth]{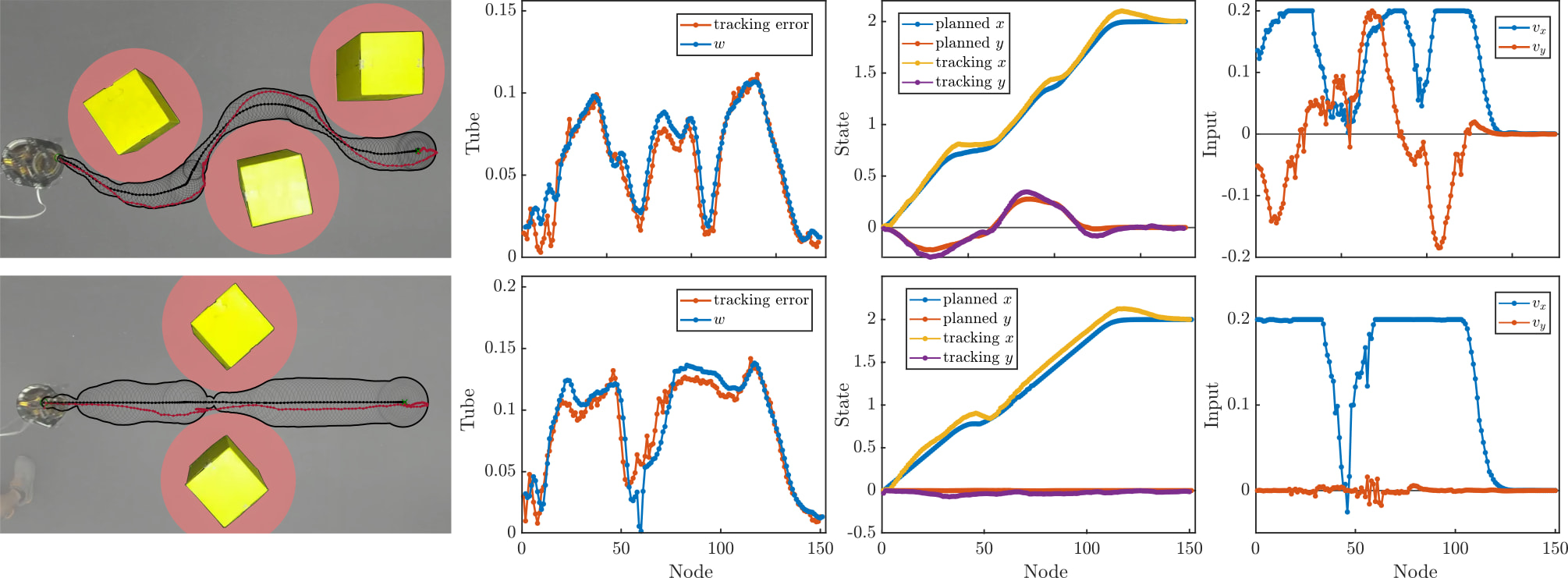}
    \caption{Implementation of Dynamic Tube MPC on the ARCHER platform. Plotted are closed loop planned trajectory and tube (white), along with tracking performance (red), along with tube, state, and input trajectories over time. Note specifically how the planning model slows down when in narrow corridors between obstacles to improve tracking.}
    \label{fig:hardware}
    \vspace{-6mm}
\end{figure*}

\section{Application of Dynamic Tube MPC to ARCHER}
We deploy our Dynamic Tube MPC method on the ARCHER platform to generate collision free trajectories in cluttered environments, in real time. 

\subsection{Learning Tube Dynamics} \label{sec:application_tube_learning}
We follow the tube learning process outlined in \cref{sub:tube_learning}, simulating the hopper with 8192 units in parallel in IsaacGym \cite{isaacgym}. We collect a dataset consisting of 409,600 trajectories, each of length 20 seconds, under random inputs to the planning model. To avoid excessive distribution shift when deploying the learned tube in MPC, we did significant work to shape the random inputs such that the planning trajectories share characteristics with the solutions from MPC; this is detailed in the code \cite{code}. Additionally, to facilitate transfer onto the hardware system, we implement domain randomization, inspired by results in RL \cite{Rudin2021, Li2024}. The ARCHER platform's tracking performance is extremely sensitive to the location of the center of mass, so we randomize this parameter within 5mm, as well as several other physical parameters of the robot (stiffness, damping, inertia, ect.). The domain randomization makes the tubes more conservative, improving the transfer to hardware. We choose a history of $H=25$, a horizon of $N=25$, and an error quantile of $\alpha = 0.9$. 
Training is conducted in PyTorch \cite{paszke2017automatic}, using neural networks with two layers of 128 nodes, and Softplus ($\beta = 5$) activations, as these result in smooth networks which are easier than ReLU to optimize through in MPC. We choose a recursive tube representation; while the one-shot tube representation was more accurate, empirically the recursive tube yielded much smoother tube predictions, and was easier to optimize through, leading to more desirable closed loop behavior in the Dynamic Tube MPC. When training the recursive models, we stop gradients on the $w$ components of $\Tilde{e}$ from \eqref{eqn:rec_tube}, as these terms depend on $\b \theta$; this stabilizes training by localizing gradient information. 

\subsection{Dynamic Tube MPC}
Due to optimizing through the neural network constraint \eqref{eqn:tube_tube}, some extra effort is required to obtain smooth solutions at real-time rates to the dynamic tube MPC problem. For our nominal problem \eqref{eqn:nom_mpc}, we choose quadratic costs on the state and input. Defining the distance to the goal as $\b e_k = \b z_k - \b z_g$, with $q,r,q_f > 0$, the cost functional is:
\begin{equation}
    J(\b z_{(\cdot)}, \b v_{(\cdot)}) = q_f \b e_k^\top \b e_k + \sum_{j=0}^{N-1}q\b e_k^\top \b e_k + r \b v_k ^\top \b v_k 
\end{equation}
In order to improve solutions to the Dynamic Tube MPC problem, \eqref{eqn:tube_mpc}, we first solve the nominal problem. Let $\bar{\b z}_k, \bar{\b v}_k$ be the solution to this problem; then the cost function for the full problem is defined as distance to the nominal problem solution. Additionally, to incentivize smooth inputs, we place a penalty on the input rate; putting these together, we have:
\begin{subequations}
 \begin{align}
    \b e_{\b z,k} &= \b z_k - \bar{\b z}_k \label{eqn:state_dev}\\
    \b e_{\b v,k} &= \b v_k - \bar{\b v}_k \label{eqn:input_dev}\\
    \b \delta_{\b v,k} &= \b v_k - \b v_{k-1} \label{eqn:input_rate}\\
    J(\b z_{(\cdot)}, \b v_{(\cdot)}) &= \sum_{k=0}^{N-1} q\b e_{\b z,k}^\top\b e_{\b z,k} + r\b e_{\b v,k} ^\top\b e_{\b v,k} \\&\phantom{=} + q_f\b e_{\b z,N}^\top\b e_{\b z,N}+\sum_{k=1}^{N-1}r_r\b \delta_{\b v,k} ^\top\b \delta_{\b v,k} \nonumber
\end{align}   
\end{subequations}
where \eqref{eqn:state_dev} and \eqref{eqn:input_dev} compute deviation from the nominal solution, and \eqref{eqn:input_rate} penalizes the input rate.

The MPC problem is solved in CasADi \cite{Andersson2019} via SNOPT \cite{gill2005snopt}, with the neural network constraint implemented via L4CasADi \cite{salzmann2023neural, salzmann2024l4casadi}. We successfully deploy our algorithm on the ARCHER hardware in a number of experimental trials. To achieve real time rates, we allow the first Dynamic Tube MPC solve to run to convergence, and then limit SNOPT to four SQP iterations per solve. Under these conditions, the Dynamic Tube MPC solves at 10Hz, which matches the $dt=0.1$ of the planning model.
% State estimation and low level control on the hopper operate at 240Hz, limited by OptiTrack measurements for position states. 

\subsection{Results}
\cref{fig:hardware} shows two experimental trials. In the bottom experiment, ARCHER is asked to navigate a straight line through a tight gap; we see the planning model operate at maximum speed, then slow to near zero speed which traversing the narrow gap, to ensure precise tracking, and finally speeding up again once free of the obstacles. In the top, similar behavior is observed, where high speeds are used when the robot is not confined between obstacles, and speeds drop in the narrow corridors.
Critically, none of these maneuvers would have been possible with a classic Robust Tube MPC implementation, without setting the velocity limit $\bar{v}$ on the planning model very low.

\subsection{Limitations}
The main limitation of this work is that the training data and evaluation data come from different distributions, as training data is random. We carefully craft our randomized training trajectories to minimize effects of this distribution shift. This could be dramatically improved by solving planning problems in the training data collection, similar to \cite{Fan2020}, using a parallelizable planning method, such as \cite{pezzato2023samplingbased}.

\section{Conclusion}

We proposed a method leveraging massively parallel simulation to learn a dynamic representation of a tracking error tube. We demonstrate that including a tracking error history in the tube dynamics dramatically improves prediction accuracy. We formulated a dynamic tube MPC, which optimized trajectories of the planning model such that the resulting dynamic tube lies in the free space, allowing real-time trade-offs between performant and conservative behaviors to ensure safety. We deploy this method on the ARCHER platform, achieving agile and collision free navigation of cluttered environments. 

%%%%%%%%%%%%%%%%%%%%%%%%%%%%%%%%%%%%%%%%%%%%%%%%%%%%%%%%%%%%%%%%%%%%%%%%%%%%%%%%

\newpage
\bibliographystyle{IEEEtran}
\balance
\bibliography{main.bib}

\end{document}